\documentclass[runningheads]{llncs}
\usepackage[T1]{fontenc}
\usepackage{glossaries-extra}
\usepackage{graphicx}

\begin{document}
\title{Falcon 7b for Software Mention Detection in Scholarly Documents}

\author{AmeerAli Khan\inst{1} \and
Qusai Ramadan\inst{3}\orcidID{0000-0001-8159-918X} \and
Cong Yang\inst{4}\orcidID{0000-0002-8314-0935} \and 
Zeyd Boukhers\inst{1,2}\orcidID{0000-0001-9778-9164}}
\authorrunning{Khan et al.}
%
\institute{Fraunhofer Institute for Applied Information Technology, Germany\\
\email{\{ameerali.khan,zeyd.boukhers\}@fit.fraunhofer.de}\\ \and
University Hospital of Cologne, Germany\\ \and
University of Koblenz, Germany\\
\email{qramadan@uni-koblenz.de}\\ \and
School of Future Science and Engineering, Soochow University, China \\
\email{cong.yang@suda.edu.cn}}

\maketitle              
\begin{abstract}
This paper aims to tackle the challenge posed by the increasing integration of software tools in research across various disciplines by investigating the application of Falcon-7b for the detection and classification of software mentions within scholarly texts. Specifically, the study focuses on solving Subtask I of the Software Mention Detection in Scholarly Publications (SOMD), which entails identifying and categorizing software mentions from academic literature. Through comprehensive experimentation, the paper explores different training strategies, including a dual-classifier approach, adaptive sampling, and weighted loss scaling, to enhance detection accuracy while overcoming the complexities of class imbalance and the nuanced syntax of scholarly writing. The findings highlight the benefits of selective labelling and adaptive sampling in improving the model's performance. However, they also indicate that integrating multiple strategies does not necessarily result in cumulative improvements. This research offers insights into the effective application of large language models for specific tasks such as SOMD, underlining the importance of tailored approaches to address the unique challenges presented by academic text analysis.

\end{abstract}

\section{Introduction}
The integration of software tools into scientific research is no longer confined to engineering disciplines but has extended to all disciplines, including the humanities and social sciences. This is largely driven by the escalating need to process and analyze data across various domains. Consequently, an extensive mention of these tools in the scientific articles. Furthermore, the volume of published scholarly articles continues to grow year after year, which makes overcoming the challenge of managing this vast repository of knowledge more important than ever before. In response, numerous projects and initiatives such as EOSC\footnote{\url{https://eosc-portal.eu/}} and NFDI\footnote{\url{https://www.nfdi.de/}} have been launched to organize research outputs, including articles, software mentions, and datasets, in a manner that adheres to the FAIR principles \cite{Wilkinson2016} (i.e.Findable, Accessible, Interoperable, and Reusable) to enhance the overall integrity and reproducibility of scientific research.

The absence of a standardized mechanism for authors to accurately cite and reference tools and datasets in their scholarly articles necessitates the extraction of such mentions post-publication. However, this task is challenging due to the unstructured nature of these articles. This underscores the importance of developing sophisticated and reliable methodologies capable of automatically and accurately recognizing mentions of these tools. This paper addresses this challenge by concentrating on a key aspect of it: "\emph{Subtask I of the Software Mention Detection in Scholarly Publications (SOMD)}." \footnote{\url{https://nfdi4ds.github.io/nslp2024/docs/somd\_shared\_task.html}} The aim is to explore new methodologies and models of software mention detection processes in academic literature. 

Although the challenge of detecting software mentions has been directly addressed using the KG approach \cite{10.1007/978-3-030-49461-2_16,10.1145/3459637.3482017}, a lot of approaches that solve the general problem have been proposed to solve the general problem of extracting information from scholarly data, including Named Entity Recognition~\cite{Singh_2023_n}, Metadata Extraction~\cite{boukhers2021mexpub,guo2023investigations,ahmed2022metadata} and Reference Extraction and Segmentation~\cite{boukhers2019end,rizvi2020hybrid,cuellar22automatic}. The proven capabilities of large language models (LLMs) to understand and generate human-like text offer new possibilities. Among the advanced models, the LLM such as the Falcon-7b model stands out due to its decoder-only architecture and the breadth of its training data. Although it was initially designed for generative tasks, the scale of architecture and the breadth of its training data present an exciting opportunity for exploring its potential for application in specialized areas of NLP, such as the detection and classification of software mentioned in academic publications. 

This paper investigates the effectiveness of the Falcon-7b model on the task of software mention detection in the scientific literature by adapting it to the task. The choice of model is arbitrary without assuming it has a specific advantage over similar models like LLAMA 2 \cite{touvron2023llama} or GPT3.5 \cite{brown2020language}. This choice allows us to explore how well such LLM models perform on a specialized task, specifically, we leverage the model's extensive training on a wide array of texts and its advanced understanding of linguistic nuances. This effectiveness is critically assessed through comprehensive experiments, encompassing a variety of evaluation metrics to ensure a balanced analysis of the model's applicability to the task at hand.

The structure of this paper is laid out as follows: Section~\ref{sec:rel} provides a review of the relevant literature. In Section~\ref{sec:met}, we delve into the methodology of this study, outlining the approaches and techniques utilized. Section~\ref{sec:eva} presents the experimental setup, alongside the results achieved from our investigations. Finally, Section~\ref{sec:con} concludes the paper, summarizing our findings and offering insights into potential avenues for future research.

\section{Related Work}
\label{sec:rel}
As there are not so many efforts to detect software mention using LLM, we review in this section approaches on named-entity recognition (NER), where we consider that software mention detection is a use case of this problem. This review is structured in three categories: 

\subsection{Rule-based and Classical Machine Learning Approaches}
Named entity recognition has been at the core of machine learning research for decades due to its importance in a variety of applications. However, the task is challenging and thus many approaches have been proposed to tackle the problem in specific scenarios. \cite{alfred2014malay}  introduces a rule-based NER framework designed for the Malay language, to improve the retrieval process of articles. It addresses the challenges faced by NER processes in languages with morphological differences and the lack of existing systems for the Malay language. By analyzing the domain of studies and the specific linguistic features of Malay, this framework accurately classifies named entities such as people, organizations, and locations. 
 \cite{morwal2012named} proposes to address the limitations of traditional rule-based approaches by using Hidden Markov Models (HMM) for NER tasks in Indian Language beyond domain-specific applications. \cite{veerasekharreddy_named_2022} proposes an approach using the Conditional Random Field (CRF) and Active Learning (AL) algorithm, where the training process of the CRF classifier is repeated until the model stabilizes. This resulted in an efficient and cost-effective outcome. \cite{gholamidastgerdi_named_2022} uses the Beam search algorithm to detect named entities in the Persian language by segmenting text into suitable and unsuitable expressions for the named entities and then applying dynamic external knowledge to recognize the emerging named entity. 

\subsection{Deep Learning-based Approaches}
Deep Neural Networks have proven significant performance over classical machine learning approaches on different tasks, including NLP, computer vision and others. Consequently, they have been used to address NER tasks as well~\cite{kesim_named_2023,li_advance_2022}. \cite{zhang2023ner} proposes a NER framework called E-NER that uses evidential deep learning (EDL) to explicitly model predictive uncertainty for named entity recognition (NER) tasks. It addresses the challenges of sparse entities and out-of-vocabulary (OOV) entities in NER by introducing uncertainty-guided loss terms and training strategies. E-NER achieves accurate uncertainty estimation, better OOV/OOD detection performance, and improved generalization ability on OOV entities compared to state-of-the-art baselines. \cite{liu2019named} introduces a novel approach to Clinical NER for de-identifying sensitive health information in clinical texts by developing a Capsule-LSTM network that leverages the strengths of capsule networks for capturing complex data relationships and LSTM networks for understanding sequential data. \cite{kong2021incorporating} the All CNN (ACNN) model for Chinese clinical NER that employs CNN enhanced by an attention mechanism, sidestepping the inefficiencies of traditional LSTM models. By leveraging multi-level CNN layers with various kernel sizes and a residual structure, ACNN adeptly captures context information across different scales, addressing the challenges posed by the complex grammar and terminology of Chinese clinical texts. \cite{cho2020combinatorial} proposes to combine CNN and bi-LSTM architectures for biomedical NER (bioNER), aiming to efficiently handle the complexities of biomedical texts, such as variant spellings and inconsistent use of prefixes and suffixes. The proposed combinatorial feature embedding and attention mechanism for enhanced entity recognition demonstrate superior performance on benchmark datasets JNLPBA and NCBI-Disease when compared with existing methods. 

\subsection{Large Language Model-based Approaches}
Due to their power to understand and generate natural text, LLMs have been employed for many Natural Language Processing (NLP) tasks, including Named Entity recognition.  \cite{wang2023gpt} proposes GPT-NER, a method that transforms the sequence labelling task of Named Entity Recognition (NER) into a text generation task, allowing large language models (LLMs) to be easily adapted for NER. GPT-NER achieves comparable performances to fully supervised baselines on five widely adopted NER datasets and exhibits a greater ability in low-resource and few-shot setups, making it suitable for real-world NER applications with limited labelled examples. \cite{oliveira2024combining} employs Generative Pre-trained Transformer 3 (GPT-3) by OpenAI together with a weak supervisor to address the NER challenge within the legal domain, exemplified by documents from the Official Gazette of the Federal District (DODF). \cite{jin2023adversarial}  proposes an architecture that refines the adversarial example generation process for LLM using disentanglement and word attribution techniques to efficiently generate adversarial examples while maintaining semantic similarity. The experiments conducted on benchmark datasets—CoNLL-2003, Ontonotes 5.0, and MultiCoNER—demonstrated that the approach improves the F1 scores by 8\%.

\section{Method}
\label{sec:met}

This section elaborates on the methodologies applied for the Software Mention Detection in Scholarly Publications (SOMD) \cite{schindler_somesci-_2021} Subtask I, part of the Natural Scientific Language Processing and Research Knowledge Graphs (NSLP) \footnote{\url{https://nfdi4ds.github.io/nslp2024/}} 2024 workshop. Our approach is centred on token classification, addressing the unique challenges software mention recognition poses in scientific texts. 

The primary objective of the subtask is to recognize software mentions within individual sentences, further classifying them by mention type (e.g., mention, usage, creation) and software type (e.g., application, programming environment, package). The task is approached with a methodology that innovatively applies a large language model (LLM), such as the Falcon-7b \cite{almazrouei_falcon_2023} model, as the foundation for a token classification system.

The core of our methodology is The Falcon-7b model, which is a decoder-only architecture known for its performance across a wide range of NLP tasks. Despite its primary design as a generative model, we adapt Falcon-7b for token classification by appending a classification layer atop its structure. This adaptation is driven by the hypothesis that the extensive pre-training on diverse corpora, coupled with its vast parameter space, provides the Falcon-7b model with a nuanced understanding of textual context. Such capabilities can significantly enhance the model's proficiency in identifying and classifying software mentions within the complex syntax and semantics of scholarly writing.

The methodology implemented in this study is structured around several strategic training approaches, tailored to address the intrinsic challenges of the SOMD task. Initially, the task is framed as a token classification problem, where labels are assigned to each word or subtoken within a sentence to denote the presence and category of software mentions. A notable challenge arises from the tendency of transformer-based models like Falcon-7b to segment words into multiple subtokens, which complicates the direct application of labels. 

To address the challenge of labelling consistency in the presence of subtoken segmentation by transformer models, our methodology employs two distinct strategies. The first, referred to as "\textbf{Unified Labeling}" assigns the same label to all subtokens derived from a single word's segmentation, ensuring consistency across the subtoken sequence. This maintains label continuity across divisions, which facilitates coherent entity recognition despite the segmentation process. In contrast, the second strategy "\textbf{Selective Labeling}" assigns a label only to the first subtoken of a segmented word, disregarding subsequent subtokens. This method aims to minimize label redundancy and computational complexity associated with processing multiple labels for a single entity. Each approach is independently explored to determine its efficacy in addressing the challenges of subtoken label alignment in the context of the SOMD task.

A critical obstacle encountered in the SOMD task is the substantial class imbalance, where data is skewed towards non-mention ('O') labels. To address this imbalance, we implemented two distinct and independently tested strategies to recalibrate the dataset for training. In the first strategy "\textbf{Weighted Loss}", weighted loss is applied, where class weights are inversely proportional to class frequencies. however, given the substantial imbalance, original weight values became impractical for underrepresented classes. To resolve this, we scaled the weights within a sensible range, setting minimums and maximums (1, and thresholds of 25, 50, 100, and 200), thereby enabling the nuanced training of models across various weight configurations to explore their efficacy in balancing classification performance. The second strategy "\textbf{Adaptive Sampling}" strategically segments the dataset into over-represented (all 'O' tokens) and under-represented (at least one non-'O' tokens) categories. This involves oversampling the under-represented data by a factor of 2 and undersampling the over-represented data to sizes equal to multiples (1, 1.5, 3) of the oversampled data volume. Adaptive Sampling aims to achieve a more balanced class distribution, enhancing the model’s capacity to learn from a representative spectrum of the dataset. Each strategy's independent evaluation provides insights into its effectiveness in addressing dataset imbalance challenges in the SOMD task.

Additionally, recognizing the multifaceted nature of software mentions, separate token classifiers are developed for identifying software types and the mentioned types. We call this strategy "\textbf{Dual-Classifier}". It enables more precise label application by separating the task into two distinct classification problems, by distinguishing between software and mention types.  This setup is designed to explore whether such a nuanced approach can effectively capture the diverse nature of software mentions, offering a potentially more sophisticated mechanism for their identification and classification.

To ensure the effectiveness and generalizability of the model, the pre-defined training dataset was utilized to develop and refine the model's capabilities.  For evaluating the model's performance in accurately identifying and classifying software mentions, the test dataset provided for Subtask II was employed. With this, we aim to conduct a comprehensive assessment of the model's ability to generalize across various scenarios and text variations found in scholarly publications, thereby validating the model's applicability and effectiveness in real-world tasks.

Following the comprehensive methodologies delineated for the Software Mention Detection (SOMD) task, our evaluation process is designed to rigorously assess the effectiveness of our approaches, namely the "Unified Labeling," "Selective Labeling," "Weighted Loss," "Adaptive Sampling," and the "Dual-Classifier" Approach. Central to our evaluation is the F1-Score, focusing on exact matches, which serves as a critical metric to quantify the precision and recall of our models in accurately identifying and classifying software mentions. Adherence to the IOB2 format for our submission files ensures our alignment with the standardized training labels, facilitating direct comparison of our model's performance against established benchmarks.

\section{Experimental Results}
\label{sec:eva}

\subsection{Results}
To evaluate our method on Software Mention Detection in Scholarly Publications (SOMD) Subtask I, we explored various settings centred around the Falcon-7b model. The experiments are conducted using the Hugging Face Transformers library\cite{wolf_huggingfaces_2020}, with PyTorch \cite{NEURIPS2019_9015} as the computational backend. Model training and evaluation were performed on a high-performance computing cluster equipped with NVIDIA A100 GPUs. To maintain consistency across all experiments conducted in this study, we standardized our training hyperparameters. This ensures that any observed differences in model performance are attributable to the variations in model architecture, data preprocessing, or other experimental conditions, rather than inconsistencies in training configurations. The hyperparameters used throughout our experiments are summarized in Table~\ref{table:hyperparameters} (Appendix~\ref{sec:app}).

The focal point of our evaluation is the F1 score, which balances precision and recall and offers a comprehensive measure of model performance. The evaluation result on the test dataset is illustrated in Table~\ref{table:evaluation_metrics}. These results are submitted to the shared task platform under the username: ``\textbf{fddaFIT}''.

 \begin{table}[h]
\centering
\caption{Evaluation metrics for each experimental approach.}
\begin{tabular}{lccc}
\hline
\textbf{Method} & \textbf{Precision} & \textbf{Recall} & \textbf{F1 Score} \\
\hline
Unified Labeling & 0.6769 & 0.4718 & 0.5561 \\
Selective Labeling & 0.7563 & 0.5243 & 0.6193 \\
Adaptive Sampling multiples@1 & 0.6496 & 0.4932 & 0.5607 \\
Adaptive Sampling multiples@1.5 & 0.7480 & \textbf{0.5417} & \textbf{0.6284} \\
Adaptive Sampling multiples@3 & 0.7213 & 0.4874 & 0.5817 \\
Weighted loss scaled@25 & 0.7559 & 0.4990 & 0.6012 \\
Weighted loss scaled@50 & 0.7211 & 0.4971 & 0.5885 \\
Weighted loss scaled@100 & 0.7195 & 0.4932 & 0.5853 \\
Weighted loss scaled@200 & 0.7110 & 0.4874 & 0.5783 \\
Dual-Classifier & \textbf{0.7602} & 0.5048 & 0.6068 \\
Dual-Classifier + Adaptive Sampling multiples@1.5 & \textbf{0.7602} & 0.5048 & 0.6068 \\
\hline
\end{tabular}
\label{table:evaluation_metrics}
\vspace{-0.7 cm}
\end{table}

In comparison between the two distinct labelling strategies: "Unified Labeling" and "Selective Labeling." The outcome favoured the "Selective Labeling" strategy, which demonstrated superior precision and recall and achieved a higher F1 score of $0.6193$ compared to the $0.5561$ of "Unified Labeling." This demonstrates the effectiveness of selectively assigning labels in enhancing the model's precision and recall. This finding guided the direction of subsequent experiments, by embedding the "Selective Labeling" strategy in our methodology.

Further investigations focus on adaptive sampling and weighted loss scaling to address the notable challenge of dataset imbalance. Adaptive sampling experiments, applying various undersampling multipliers to over-represented data, demonstrated the precise impact of data distribution on model efficacy. The employment of a multiplier of 1.5 showed an enhancement of the F1 score to $0.6284$, the highest among adaptive sampling variations. This suggests an optimal balance in dataset composition, significantly contributing to model performance.

Experiments with weighted loss, adjusted across a range of maximum weights (25, 50, 100, and 200), aimed to refine the model's sensitivity to class frequencies. While experimenting with weighted loss scaling offered insights into handling class imbalance, the adjustments did not surpass the adaptive sampling's peak F1 score, with the highest recorded F1 at $0.6012$ for a scaling factor of $25$.

The "Dual-Classifier" approach introduced a bifurcated strategy for identifying software types and mention types, aiming to enrich the model's understanding and classification accuracy. Interestingly, this approach alone achieved an F1 score of 0.6068, comparable to some weighted loss scaling strategies but slightly below the best-performing adaptive sampling method. Combining the "Dual-Classifier" with "Adaptive Sampling" at a multiplier of 1.5 did not further enhance the F1 score, indicating that while each method independently contributes to addressing specific challenges in SOMD, their combined effect does not necessarily result in cumulative improvements.

The experiments demonstrate the critical role of selective labelling and adaptive sampling in enhancing F1 scores for SOMD. While weighted loss scaling and the "Dual-Classifier" approach contribute to performance improvements, the combination of strategies does not yield further enhancements. This indicates the necessity of strategic selection and implementation in model development for SOMD tasks.


In these experiments, we observed that the model effectively identifies software mentions within the complex academic text and accurately classifies entities like specific software tools or programming languages.

For example, as shown in Table~\ref{table:confusion_usage_mention} (Appendix~\ref{sec:app}), it might correctly identify "Python" as a programming language used within a research context. However, challenges arise in distinguishing between mentions of software and instances where the software is being actively used or discussed in depth. This differentiation is crucial for understanding the role of software in research, as only mentions might not signify importance or relevance to the study's outcomes. Examples of the model's output are provided in the full version\footnote{\url{https://zenodo.org/doi/10.5281/zenodo.10993039}} of this paper.


\section{Conclusion}
\label{sec:con}

In this paper, we presented the efficacy of the Falcon-7b model, a prominent Large Language Model (LLM), in tackling the nuances of Software Mention Detection (SOMD) within scholarly publications. Guided by the hypothesis that advanced LLMs could significantly improve the precision and recall for SOMD tasks due to their extensive training on diverse datasets, our study systematically explored various strategies centred around the Falcon-7b model.

The comparative analysis of "Unified Labeling" and "Selective Labeling" strategies revealed a preference for "Selective Labeling," which yielded a higher F1 score. This result underscores the incremental nature of advancements achievable with the Falcon-7b model in the context of SOMD tasks. Additionally, our experiments with adaptive sampling and weighted loss scaling aimed at addressing dataset imbalances highlighted that, despite certain improvements, the enhancements were not as substantial as anticipated. The adaptive sampling strategy, especially with a multiplier of 1.5, did indeed enhance the F1 score, to the highest obtained score. However, this observation was not sufficient to categorize the performance as extraordinary.  This finding suggests that while the Falcon-7b model and similar LLMs hold promise for NLP tasks, their application in specialized areas such as SOMD may not always yield groundbreaking results.

In conclusion, the outcomes of our study indicate that, despite the advanced capabilities of LLMs like Falcon-7b, the performance improvements in specific NLP tasks such as SOMD are modest. While LLMs offer advantages in processing and understanding complex language patterns, their effectiveness in specialized domains like software mention detection within scholarly texts does not markedly outperform more traditional approaches. This emphasizes the importance of combining LLMs with other approaches and the need to explore a broad spectrum of models and methodologies to identify the most effective solutions for specialized NLP challenges. 
\section*{Acknowledgement}
This work was funded by the FAIR Data Spaces project of the German Federal Ministry of Education and Research (BMBF)
under grant numbers FAIRDS05 and FAIRDS15 and by NFDI4DS project with grant number 460234259.

\bibliographystyle{splncs04}
\bibliography{references}

\begin{thebibliography}{10}
\providecommand{\url}[1]{\texttt{#1}}
\providecommand{\urlprefix}{URL }
\providecommand{\doi}[1]{https://doi.org/#1}

\bibitem{ahmed2022metadata}
Ahmed, R.M.W.: Metadata Extraction using Geometric and Layout Features from Research Publications. Ph.D. thesis, CAPITAL UNIVERSITY (2022)

\bibitem{alfred2014malay}
Alfred, R., Leong, L.C., On, C.K., Anthony, P.: Malay named entity recognition based on rule-based approach  (2014)

\bibitem{almazrouei_falcon_2023}
Almazrouei, E., Alobeidli, H., Alshamsi, A., Cappelli, A., Cojocaru, R., Debbah, M., Goffinet, E., Hesslow, D., Launay, J., Malartic, Q., Mazzotta, D., Noune, B., Pannier, B., Penedo, G.: The {Falcon} {Series} of {Open} {Language} {Models} (Nov 2023). \doi{10.48550/arXiv.2311.16867}, \url{http://arxiv.org/abs/2311.16867}, arXiv:2311.16867 [cs]

\bibitem{boukhers2019end}
Boukhers, Z., Ambhore, S., Staab, S.: An end-to-end approach for extracting and segmenting high-variance references from pdf documents. In: 2019 ACM/IEEE Joint Conference on Digital Libraries (JCDL). pp. 186--195. IEEE (2019)

\bibitem{boukhers2021mexpub}
Boukhers, Z., Beili, N., Hartmann, T., Goswami, P., Zafar, M.A.: Mexpub: Deep transfer learning for metadata extraction from german publications. In: 2021 ACM/IEEE Joint Conference on Digital Libraries (JCDL). pp. 250--253. IEEE (2021)

\bibitem{brown2020language}
Brown, T.B., Mann, B., Ryder, N., Subbiah, M., Kaplan, J., Dhariwal, P., Neelakantan, A., Shyam, P., Sastry, G., Askell, A., Agarwal, S., Herbert-Voss, A., Krueger, G., Henighan, T., Child, R., Ramesh, A., Ziegler, D.M., Wu, J., Winter, C., Hesse, C., Chen, M., Sigler, E., Litwin, M., Gray, S., Chess, B., Clark, J., Berner, C., McCandlish, S., Radford, A., Sutskever, I., Amodei, D.: Language models are few-shot learners (2020)

\bibitem{cho2020combinatorial}
Cho, M., Ha, J., Park, C., Park, S.: Combinatorial feature embedding based on cnn and lstm for biomedical named entity recognition. Journal of biomedical informatics  \textbf{103},  103381 (2020)

\bibitem{cuellar22automatic}
Cu{\'e}llar-Hidalgo, R., Reyes-Salgado, G., Torres-Moreno, J.M.: Automatic reference mining: Review and perspectives. TextMine'22

\bibitem{gholamidastgerdi_named_2022}
Gholami‐Dastgerdi, P., Feizi‐Derakhshi, M., Forouzandeh, A.: Named entities detection by beam search algorithm. Concurrency and Computation: Practice and Experience  \textbf{34}(27),  e7325 (Dec 2022). \doi{10.1002/cpe.7325}, \url{https://onlinelibrary.wiley.com/doi/10.1002/cpe.7325}

\bibitem{guo2023investigations}
Guo, M., Wu, F., Jiang, J., Yan, X., Chen, G., Li, W., Zhao, Y., Sun, Z.: Investigations on scientific literature meta information extraction using large language models. In: 2023 IEEE International Conference on Knowledge Graph (ICKG). pp. 249--254. IEEE (2023)

\bibitem{jin2023adversarial}
Jin, X., Vinzamuri, B., Venkatapathy, S., Ji, H., Natarajan, P.: Adversarial robustness for large language ner models using disentanglement and word attributions. In: The 2023 Conference on Empirical Methods in Natural Language Processing (2023)

\bibitem{kesim_named_2023}
Kesim, E., Deliahmetoglu, A.: Named entity recognition in resumes (Jun 2023). \doi{10.48550/arXiv.2306.13062}, \url{http://arxiv.org/abs/2306.13062}, arXiv:2306.13062 [cs]

\bibitem{kong2021incorporating}
Kong, J., Zhang, L., Jiang, M., Liu, T.: Incorporating multi-level cnn and attention mechanism for chinese clinical named entity recognition. Journal of Biomedical Informatics  \textbf{116},  103737 (2021)

\bibitem{li_advance_2022}
Li, W.: The {Advance} of {Deep} {Learning} {Based} {Named} {Entity} {Recognition}. Highlights in Science, Engineering and Technology  \textbf{12},  68--73 (Aug 2022). \doi{10.54097/hset.v12i.1368}, \url{https://drpress.org/ojs/index.php/HSET/article/view/1368}

\bibitem{liu2019named}
Liu, C., Li, J., Liu, Y., Du, J., Tang, B., Xu, R.: Named entity recognition in clinical text based on capsule-lstm for privacy protection. In: Artificial Intelligence and Mobile Services--AIMS 2019: 8th International Conference, Held as Part of the Services Conference Federation, SCF 2019, San Diego, CA, USA, June 25--30, 2019, Proceedings 8. pp. 166--178. Springer (2019)

\bibitem{morwal2012named}
Morwal, S., Jahan, N., Chopra, D.: Named entity recognition using hidden markov model (hmm). International Journal on Natural Language Computing (IJNLC) Vol  \textbf{1} (2012)

\bibitem{oliveira2024combining}
Oliveira, V., Nogueira, G., Faleiros, T., Marcacini, R.: Combining prompt-based language models and weak supervision for labeling named entity recognition on legal documents. Artificial Intelligence and Law pp. 1--21 (2024)

\bibitem{NEURIPS2019_9015}
Paszke, A., Gross, S., Massa, F., Lerer, A., Bradbury, J., Chanan, G., Killeen, T., Lin, Z., Gimelshein, N., Antiga, L., Desmaison, A., Kopf, A., Yang, E., DeVito, Z., Raison, M., Tejani, A., Chilamkurthy, S., Steiner, B., Fang, L., Bai, J., Chintala, S.: Pytorch: An imperative style, high-performance deep learning library. In: Advances in Neural Information Processing Systems 32, pp. 8024--8035. Curran Associates, Inc. (2019), \url{http://papers.neurips.cc/paper/9015-pytorch-an-imperative-style-high-performance-deep-learning-library.pdf}

\bibitem{rizvi2020hybrid}
Rizvi, S.T.R., Dengel, A., Ahmed, S.: A hybrid approach and unified framework for bibliographic reference extraction. IEEE Access  \textbf{8},  217231--217245 (2020)

\bibitem{10.1145/3459637.3482017}
Schindler, D., Bensmann, F., Dietze, S., Kr\"{u}ger, F.: Somesci- a 5 star open data gold standard knowledge graph of software mentions in scientific articles. In: Proceedings of the 30th ACM International Conference on Information \& Knowledge Management. p. 4574–4583. CIKM '21, Association for Computing Machinery, New York, NY, USA (2021). \doi{10.1145/3459637.3482017}, \url{https://doi.org/10.1145/3459637.3482017}

\bibitem{schindler_somesci-_2021}
Schindler, D., Bensmann, F., Dietze, S., Krüger, F.: {SoMeSci}- {A} 5 {Star} {Open} {Data} {Gold} {Standard} {Knowledge} {Graph} of {Software} {Mentions} in {Scientific} {Articles}. In: Proceedings of the 30th {ACM} {International} {Conference} on {Information} \& {Knowledge} {Management}. pp. 4574--4583. ACM, Virtual Event Queensland Australia (Oct 2021). \doi{10.1145/3459637.3482017}, \url{https://dl.acm.org/doi/10.1145/3459637.3482017}

\bibitem{10.1007/978-3-030-49461-2_16}
Schindler, D., Zapilko, B., Kr{\"u}ger, F.: Investigating software usage in the social sciences: A knowledge graph approach. In: Harth, A., Kirrane, S., Ngonga~Ngomo, A.C., Paulheim, H., Rula, A., Gentile, A.L., Haase, P., Cochez, M. (eds.) The Semantic Web. pp. 271--286. Springer International Publishing, Cham (2020)

\bibitem{Singh_2023_n}
Singh, A., Garg, A.: Named {Entity} {Recognition} ({NER}) and {Relation} {Extraction} in {Scientific} {Publications}. International Journal of Recent Technology and Engineering (IJRTE)  \textbf{12}(2),  110--113 (2023). \doi{10.35940/ijrte.B7846.0712223}, \url{https://www.ijrte.org/portfolio-item/B78460712223/}

\bibitem{touvron2023llama}
Touvron, H., Martin, L., Stone, K., Albert, P., Almahairi, A., Babaei, Y., Bashlykov, N., Batra, S., Bhargava, P., Bhosale, S., Bikel, D., Blecher, L., Ferrer, C.C., Chen, M., Cucurull, G., Esiobu, D., Fernandes, J., Fu, J., Fu, W., Fuller, B., Gao, C., Goswami, V., Goyal, N., Hartshorn, A., Hosseini, S., Hou, R., Inan, H., Kardas, M., Kerkez, V., Khabsa, M., Kloumann, I., Korenev, A., Koura, P.S., Lachaux, M.A., Lavril, T., Lee, J., Liskovich, D., Lu, Y., Mao, Y., Martinet, X., Mihaylov, T., Mishra, P., Molybog, I., Nie, Y., Poulton, A., Reizenstein, J., Rungta, R., Saladi, K., Schelten, A., Silva, R., Smith, E.M., Subramanian, R., Tan, X.E., Tang, B., Taylor, R., Williams, A., Kuan, J.X., Xu, P., Yan, Z., Zarov, I., Zhang, Y., Fan, A., Kambadur, M., Narang, S., Rodriguez, A., Stojnic, R., Edunov, S., Scialom, T.: Llama 2: Open foundation and fine-tuned chat models (2023)

\bibitem{veerasekharreddy_named_2022}
VeeraSekharReddy, B., Rao, K.S., Koppula, N.: Named {Entity} {Recognition} using {CRF} with {Active} {Learning} {Algorithm} in {English} {Texts}. In: 2022 6th {International} {Conference} on {Electronics}, {Communication} and {Aerospace} {Technology}. pp. 1041--1044 (Dec 2022). \doi{10.1109/ICECA55336.2022.10009592}, \url{https://ieeexplore.ieee.org/document/10009592}

\bibitem{wang2023gpt}
Wang, S., Sun, X., Li, X., Ouyang, R., Wu, F., Zhang, T., Li, J., Wang, G.: Gpt-ner: Named entity recognition via large language models. arXiv preprint arXiv:2304.10428  (2023)

\bibitem{Wilkinson2016}
Wilkinson, M.D., Dumontier, M., Aalbersberg, I.J., Appleton, G., Axton, M., Baak, A., Blomberg, N., Boiten, J.W., da~Silva~Santos, L.B., Bourne, P.E., Bouwman, J., Brookes, A.J., Clark, T., Crosas, M., Dillo, I., Dumon, O., Edmunds, S., Evelo, C.T., Finkers, R., Gonzalez-Beltran, A., Gray, A.J., Groth, P., Goble, C., Grethe, J.S., Heringa, J., ’t Hoen, P.A., Hooft, R., Kuhn, T., Kok, R., Kok, J., Lusher, S.J., Martone, M.E., Mons, A., Packer, A.L., Persson, B., Rocca-Serra, P., Roos, M., van Schaik, R., Sansone, S.A., Schultes, E., Sengstag, T., Slater, T., Strawn, G., Swertz, M.A., Thompson, M., van~der Lei, J., van Mulligen, E., Velterop, J., Waagmeester, A., Wittenburg, P., Wolstencroft, K., Zhao, J., Mons, B.: The fair guiding principles for scientific data management and stewardship. Scientific Data  \textbf{3}(1) (Mar 2016). \doi{10.1038/sdata.2016.18}, \url{http://dx.doi.org/10.1038/sdata.2016.18}

\bibitem{wolf_huggingfaces_2020}
Wolf, T., Debut, L., Sanh, V., Chaumond, J., Delangue, C., Moi, A., Cistac, P., Rault, T., Louf, R., Funtowicz, M., Davison, J., Shleifer, S., von Platen, P., Ma, C., Jernite, Y., Plu, J., Xu, C., Scao, T.L., Gugger, S., Drame, M., Lhoest, Q., Rush, A.M.: {HuggingFace}'s {Transformers}: {State}-of-the-art {Natural} {Language} {Processing} (Jul 2020). \doi{10.48550/arXiv.1910.03771}, \url{http://arxiv.org/abs/1910.03771}, arXiv:1910.03771 [cs]

\bibitem{zhang2023ner}
Zhang, Z., Hu, M., Zhao, S., Huang, M., Wang, H., Liu, L., Zhang, Z., Liu, Z., Wu, B.: E-ner: Evidential deep learning for trustworthy named entity recognition. In: Findings of the Association for Computational Linguistics: ACL 2023. pp. 1619--1634 (2023)

\end{thebibliography}
\end{document}